\documentclass[runningheads]{llncs}
\usepackage[T1]{fontenc}
\usepackage{graphicx}
\usepackage{amsmath}
\usepackage{bm}
\usepackage{booktabs}
\usepackage{hyperref}
\usepackage{tabularx}

\DeclareMathOperator*{\argmin}{arg\,min}

\newcommand{\icol}[1]{
  \left(\begin{smallmatrix}#1\end{smallmatrix}\right)%
}

\usepackage{color}

\begin{document}
\title{CINA: Conditional Implicit Neural Atlas for Spatio-Temporal Representation of Fetal Brains}
\titlerunning{CINA: Conditional Implicit Neural Atlas}
%
\author{Maik Dannecker\inst{1}
\and
Vanessa Kyriakopoulou\inst{3}
\and
Lucilio Cordero-Grande\inst{4}
\and
Anthony N. Price\inst{3}
\and
Joseph V. Hajnal\inst{3}
\and
Daniel Rueckert\inst{1, 2}}
\authorrunning{M. Dannecker et al.}
%
\institute{School of Computation, Information and Technology, and School of Medicine and Health, Technical University Munich, Germany
\email{m.dannecker@tum.de}\\
\and
Department of Computing, Imperial College London, United Kingdom 
\and
Centre for the Developing Brain, School of Biomedical Engineering and Imaging Sciences, King’s College London, United Kingdom\\
\and
Biomedical Image Technologies, ETSI Telecomunicación, Universidad Politécnica de Madrid, Spain\\
} 

\maketitle              
%
\begin{abstract}
We introduce a conditional implicit neural atlas (CINA) for spatio-temporal atlas generation from Magnetic Resonance Images (MRI) of the neurotypical and pathological fetal brain, that is fully independent of affine or non-rigid registration. During training, CINA learns a general representation of the fetal brain and encodes subject specific information into latent code. After training, CINA can construct a faithful atlas with tissue probability maps of the fetal brain for any gestational age (GA) and anatomical variation covered within the training domain. Thus, CINA is competent to represent both, neurotypical and pathological brains. Furthermore, a trained CINA model can be fit to brain MRI of unseen subjects via test-time optimization of the latent code. CINA can then produce probabilistic tissue maps tailored to a particular subject. We evaluate our method on a total of 198 T2 weighted MRI of normal and abnormal fetal brains from the dHCP and FeTA datasets. We demonstrate CINA's capability to represent a fetal brain atlas that can be flexibly conditioned on GA and on anatomical variations like ventricular volume or degree of cortical folding, making it a suitable tool for modeling both neurotypical and pathological brains. We quantify the fidelity of our atlas by means of tissue segmentation and age prediction and compare it to an established baseline. CINA demonstrates superior accuracy for neurotypical brains and pathological brains with ventriculomegaly. Moreover, CINA scores a mean absolute error of 0.23 weeks in fetal brain age prediction, further confirming an accurate representation of fetal brain development.

\end{abstract}
\section{Introduction}
The spatio-temporal analysis of fetal brain structures from MRI constitutes a critical cornerstone in understanding normal and abnormal growth patterns of the early developing brain. Fetal brain atlases serve as important instrument for precise characterization of neural development, facilitating numerous applications in medical image analysis including image segmentation and shape analysis. As fetal brains experience rapid development throughout pregnancy, they demand an atlas with high resolution in space and time to be adequately modeled across gestational age (GA). Existent work focuses on classical registration based approaches to map a set of brains of similar GA to a common reference \cite{Gholipour2017,Habas2010,Kuklisova2011,Makropoulus2016,Schuh2014,Serag2012,Serag2012_fetal}.
These approaches, however, come with notable limitations. They rely on compute intensive, non-rigid registration schemes and are restricted to discrete time points and fixed spatial resolution. More critically, the atlases typically only represent a normative brain and fail to adequately represent the development and characteristics of brains in the presence of pathologies \cite{FeTA2021}. Studying brains with distinct pathologies requires an atlas based on brains exhibiting similar conditions. However, datasets for such specific pathologies are usually limited. Hence, established methods struggle to generate a temporal unbiased atlas of the fetal brain that faithfully represents pathological growth patterns.

\noindent\textbf{Contribution.} We propose CINA, a conditional implicit neural atlas based on Implicit Neural Representation (INR) that learns a continuous, spatio-temporal representation of the fetal brain with corresponding tissue probability maps for brain segmentation. Our method facilitates flexible atlas generation of arbitrary resolution in time and space. Additionally, CINA can condition on specific anatomical features during training and therefore interpolate within the anatomical feature space for atlas generation. E.g., in training, CINA conditions on brain Gyrification Index or ventricular volume and can then create brain atlases with varying degrees of cortical folding or with ventricles of different sizes, respectively. We evaluate CINA on a total of 198 fetal brain MRI, including neurotypical brains from the dHCP dataset \cite{dhcp_data} and pathological brains with ventriculomegaly (VM) from the FeTA dataset \cite{FeTA2021}. CINA provides accurate tissue probability maps, especially for pathological brains with VM, and demonstrates impressive age prediction accuracy with a mean absolute error of 0.23 weeks for neurotypical brains.

\noindent\textbf{Related Work.} Initial work employed pairwise affine registration with kernel regression across age to construct an atlas from MRI of preterm neonates \cite{Kuklisova2011}. Follow up work extended the registration model to non-rigid deformations using free-form deformation models \cite{Makropoulus2016,FFDRueckert,Serag2012} or diffeomorphic registration \cite{Schuh2014}. First work on a fetal brain atlas used polynomial fitting and non-rigid deformation to construct a spatio-temporal atlas of 20 neurotypical fetal brains \cite{Habas2010}. This was followed by an FFD approach from \cite{Serag2012_fetal} constructing a fetal brain atlas from MRI of 80 healthy fetuses spanning a GA of 22 to 38 weeks. in 2017, Gholipour et al. \cite{Gholipour2017} constructed a spatio-temporal MRI atlas with corresponding brain tissue segmentation from 81 healthy fetuses of 21-37 weeks GA. However, these models are built from neurotypical populations and therefore inadequate to model pathological growth patters in fetal brains. 

Recently, neural network based method for conditional atlas creation \cite{cheng2020unbiased,Dalca2019,Neel2021} can provide a more flexible brain representation. Implicit Neural Representations (INRs) promise a further direction for conditional spatio-temporal atlas construction. INRs model data with a multilayer perceptron (MLP) as continuous functions of coordinates and comprise a resolution agnostic representation of the data \cite{Mildenhall2021,Mueller2022,Wire,Siren,TancikFourier}. Typically, an INR is (over-)fit on a specific data instance without exploiting any task-specific knowledge as prior, limiting the representation. However, multiple strategies to fit an INR to an entire dataset across a domain exist \cite{FunctaSpatial2023,Functa2022,Mehta,MeschederOccup,Park2019,SitzmanConditioned}. Here, the INR represents domain knowledge, whereas instance specific information is encoded in a instance specific learnable latent code. During training, the latent code can either be directly fed to the INR by concatenation to the input coordinates \cite{MeschederOccup,Park2019} or mapped to the weights of the INR using a hypernetwork \cite{Siren,SitzmanConditioned}. Other work proposes to linearly map the latent code to shift and scale parameters which then modulate the output of the INR layers~\cite{Functa2022,Mehta}. This approach demonstrates enhanced reconstruction quality during inference and fast training. The latent code itself can be optimized through an encoder or auto-decoder approach \cite{Park2019,MetaSDF,Siren}. In the auto-decoder approach, latent code is randomly initialized, usually Gaussian distributed close to zero, and optimized during training by backpropagation through the INR, end-to-end \cite{Park2019}. During inference on a unseen signal instance, a new initialized latent is fit to the signal while freezing the INR. The auto-decoder setup is fully resolution agnostic and fast to train, but slower for inference due to the latent code optimization.     

\section{Methodology}
\begin{figure}[t!]
\includegraphics[width=\textwidth]{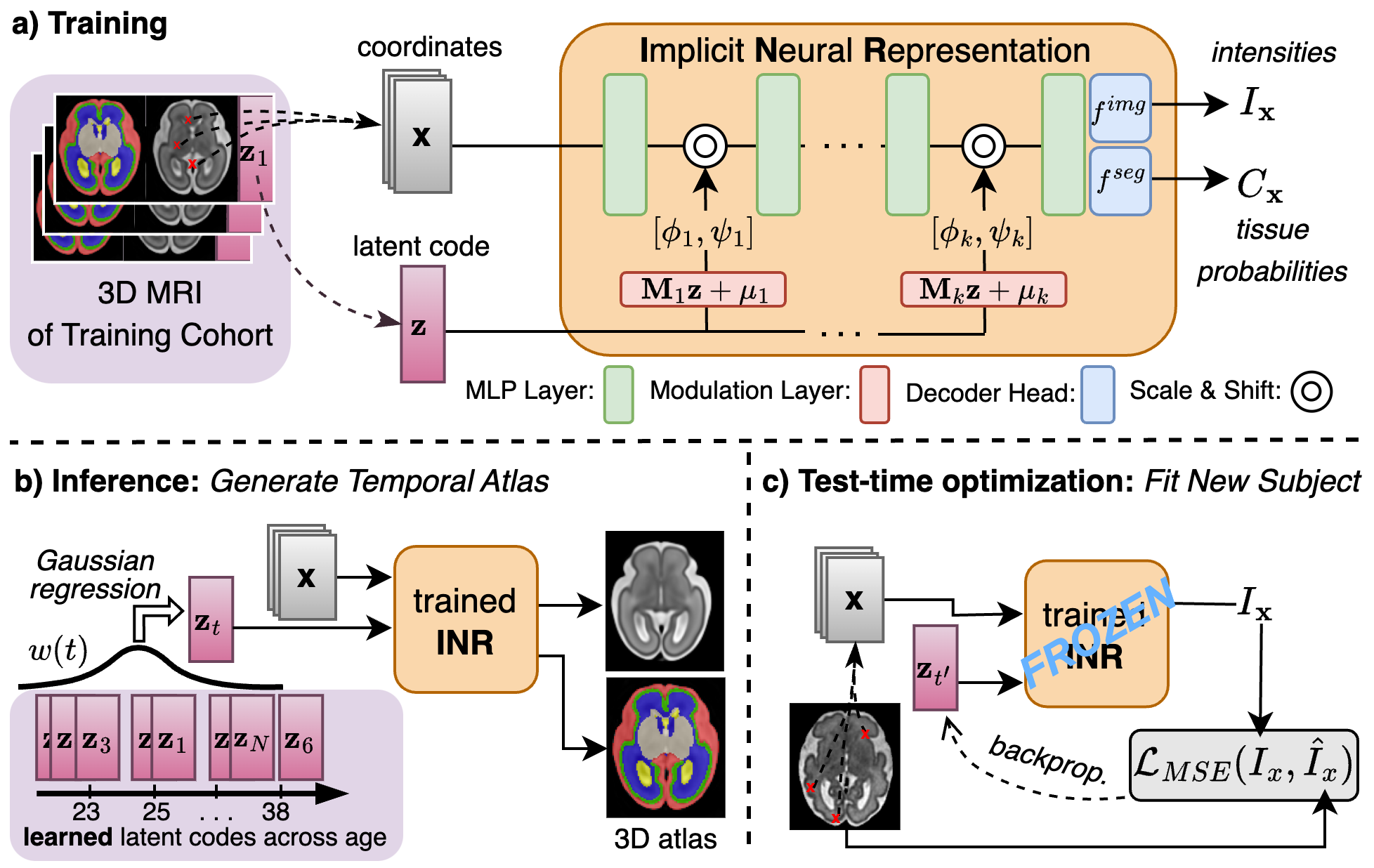}
\caption{CINA comprises three phases. \textbf{a)} In the training phase, CINA learns to reconstruct 3D MRI with tissue segmentations of $N$ subjects. Each subject is assigned its unique \textit{trainable} latent code $\textbf{z}$. Thus, the INR learns general domain knowledge while subject specific information is represented in $\textbf{z}$. \textbf{b)} CINA has learned a spatio-temporal representation of the fetal brain. To generate an atlas for time $t$, we regress a new $\textbf{z}_t$ from the trained latents $\{\textbf{z}_i\}_{i=1}^N$. Then, a single forward pass yields the atlas with tissue probability maps. \textbf{c)} To fit the atlas to a new subject of age $t'$, we perform test-time optimization on the MRI intensities, learning a new latent $\textbf{z}_t'$ while freezing the INR. Finally, we get the atlas and probability maps, tailored to the subject, via a forward pass while conditioning on $\textbf{z}_t'$.}         
\label{architecture}
\end{figure}
CINA is modeled as an auto-decoder approach, consisting of a fully-connected network with sinusoidal activation functions following \cite{Siren}. To condition on latent code, we employ modulated layers as in \cite{Functa2022,Mehta}. The latent code $\textbf{z}$ is mapped to scale $(\bm{\phi})$ and shift $(\bm{\psi})$ parameters through a linear modulation layer, i.e., \(\icol{\bm{\phi}\\\bm{\psi}} = \textbf{Mz}+\bm{\mu}\). The output of a modulated layer therefore equates to
\(\textnormal{sin}\left(\left(\omega_0 \cdot \bm{\phi} \cdot \textbf{Wx}+\textbf{b}\right) + \bm{\psi}\right)\)
where $\omega_0$ denotes a multiplicative term to better fit high frequency signals \cite{Siren}. Hence, different to \cite{Functa2022}, we do not want the modulation shift $\bm{\psi}$ to be scaled by $\omega_0$, since the latent code represents temporal changes of the brain and therefore should predominantly capture low-frequency features to enforce a smooth latent representation. Similar to \cite{Park2019}, each subject of the training cohort has its unique latent code $\textbf{z}$ which is initialized close to zero as \(\textbf{z} \sim \mathcal{N}(0, 10^{-2})\). This helps the model to learn a compact representation of \textbf{z}, facilitating smooth interpolation within the latent code. 

\noindent \textbf{Training.}
Fig.~\ref{architecture} a) depicts the training process. We maximize the joint log posterior over all $N$ training subjects, optimizing the respective latent codes $\{\textbf{z}_i\}_{i=1}^N$ and the model parameters $\theta$:
\begin{equation}
    \argmin_{\theta, \{\textbf{z}_i\}_{i=1}^N} = \sum_{i=1}^N\left(
    \sum_{\textbf{x}\in\textbf{X}_i}\mathcal{L}_{MSE}\left(
    f_{\theta}^{img}\left(\textbf{x}, \textbf{z}_i\right), \hat{I}_\textbf{x}\right)
    + \mathcal{L}_{CE}\left(f_{\theta}^{seg}\left(\textbf{x}, \textbf{z}_i\right), \hat{C}_\textbf{x}\right)
    \right)
\end{equation}
Here $\hat{I}_\textbf{x}$ and $\hat{C}_\textbf{x}$ denote the groundtruth intensity and tissue label of voxel \textbf{x}, $f^{img}$ and $f^{seg}$ denote the decoder heads of the INR, and $\mathcal{L}_{MSE}$ and $\mathcal{L}_{CE}$ denote the mean-squared error and the cross entropy loss. We found that no explicit regularization for $\textbf{z}$ is necessary during training to enforce a compact latent representation.

\noindent\textbf{Atlas Construction.}
After training, the INR has learned a general representation of the target domain, i.e., the fetal brain, having pushed all individual information into the subject specific latent codes $\{\textbf{z}_i\}_{i=1}^N$. This is in particular the case for GA and confirmed by Fig.~\ref{fig:age_correlation}. Illustrated in Fig.~\ref{architecture} b), this allows us to generate a brain atlas for any given time point $t$ by using a time regressed latent $\textbf{z}_t$ defined as
\begin{equation}
    \textbf{z}_t = \sum_{i=1}^{N}w(t, t_i)\textbf{z}_i,
\label{eq:mean_latent}
\end{equation}
where $w$ represents a Gaussian kernel with mean $t$ and a standard-deviation $\sigma$, empirically set to $\sigma=0.35$ for all experiments. Finally, by conditioning on the regressed latent $\textbf{z}_t$, we get the atlas with intensities $I_\textbf{x}$ and tissue probability maps $C_\textbf{X}$ for voxels $\textbf{X}$ from the trained INR via a forward pass:
\begin{equation}
\label{eq:forward_pass}
\textnormal{INR}_\theta(\textbf{X}|\textbf{z}_t) = \left(f_{\theta}^{img}\left(\textbf{X}, \textbf{z}_t\right), f_{\theta}^{seg}\left(\textbf{X}, \textbf{z}_t\right)\right) = (I_{\textbf{X}}, C_{\textbf{X}}).
\end{equation}

\noindent \textbf{Atlas matching to new subjects.}
\label{inference}
To match the atlas to a new subject of gestational age $t'$, we optimize a new latent code $\textbf{z}_t'$ to the MRI intensities $\hat{I}(\textbf{X})$, while freezing the INR parameters $\theta$, minimizing
\begin{equation}
    \textbf{z}_{t'} = \argmin_{\textbf{z}} \sum_{\textbf{x}\in \textbf{X}} \mathcal{L}\left(f_\theta^{img}\left(\textbf{X}, \textbf{z}\right), \hat{I}\left(\textbf{x}\right)\right) + \frac{1}{\sigma^2}\Vert\textbf{z}\Vert_2^2,
\end{equation} where $\textbf{z}_{t'}$ is initialized according to Eq.~\ref{eq:mean_latent} or as $\textbf{z}_{t'} \sim \mathcal{N}(0, 10^{-2})$ should the subject's age be unknown. To enforce a compact representation of $\textbf{z}$ and to prevent overfitting, we apply $L_2$ regularization for inference. We can also use secondary information for latent code optimization, e.g., tissue segmentation, to gain a richer representation. In this work, however, we only use the scan intensities to demonstrate the methods capability to infer accurate tissue probability maps from the optimized latent code. We generate these probability maps via Eq.~\ref{eq:forward_pass}.

\subsection{Conditioning on Anatomical Characteristics}
\label{anat_conditioning}
CINA implicitly conditions on the patterns of anatomical variations that are present in the training population and encodes these variations in the subject specific latent code $\textbf{z}$. We can also force CINA to explicitly condition on specific brain anatomy to create an atlas tailored to a sub-population, e.g. subjects with ventriculomegaly (VM), by quantifying and incorporating the desired brain anatomy as additional dimension into $z$. We demonstrate the procedure on two examples: Volume of the lateral ventricles (LV) and the degree of cortical folding. 

\noindent\textbf{Lateral Ventricular Volume.}
Conditioning on the LV volume is particularly helpful to adequately model brains with VM. Whereas diagnosis of VM usually involves the width of the atrial diameter, there also exists a linear relationship with the LV volume \cite{Ma_VM_2019}. As the FeTA dataset provides segmentations of the LV but not the width of the atrial diameter, we decided to condition on the volume of the LV which is easily computed from the segmentation.  

\noindent\textbf{Cortical Folding.}
We further demonstrate the effectiveness of this approach on a more complex shape by conditioning on the cortex, specifically on the degree of cortical folding.
We quantify the degree of folding with the Gyrification Index \cite{zilles1988human} using the cortical gray matter segmentation provided with FeTA.
\section{Experiments and Results}
\noindent\textbf{Datasets.}
In this study we use the two fetal brain datasets, dHCP \cite{dhcp_data} and FeTA \cite{FeTA2021}. From the dHCP dataset, we use T2 weighted 3D MRI of 138 fetal brains showing no abnormalities. Each 3D volume comes with an isotropic resolution of 0.5 mm and a quality controlled FreeSurfer \cite{Freesurfer2021,dHCP_pipe} segmentation of 16 anatomical regions. The age range at the time of scan was between 22 and 38 weeks gestational age (GA). Further details on acquisition and pre-processing parameters can be found under \cite{dhcp_data}. 

The FeTA dataset contains T2 weighted 3D MRI of 80 neurotypical and pathological fetal brains. All scans are of 0.5 mm isotropic resolution and come with ground truth segmentations for 7 anatomical regions annotated by experts. For more details on acquisition and annotation protocol, we refer the reader to \cite{FeTA2021}. Of the 80 scans we discarded 20 scans due to insufficient quality leaving 30 normal and 30 abnormal cases within a age range of 21 to 36 weeks GA. Fig.~\ref{fig:histogram} of the supplementary material shows the detailed distributions of the two datasets.   

\noindent\textbf{Preprocessing.}
We rigidly aligned all brain volumes to MNI space. Next, we harmonized the label maps of dHCP and FeTA data by merging smaller regions so that we ended up with the six anatomical regions: Cerebrospinal fluid (CSF), cortical Gray Matter(cGM), White Matter (WM), Lateral Ventricles (LV), Cerebellum (CB), and Brainstem (BS). 

\noindent\textbf{Training Setup.}
We trained two CINA models. One on a population of 128 healthy fetal subjects from the dHCP dataset \cite{dhcp_data} and one on a population of 25 healthy and 25 pathological subjects from the FeTA dataset \cite{FeTA2021}. We configured CINA with 5 hidden layers of size 512, modulation layers of size 330 after the first, third, and fifth hidden layer, and latent code size of $\textbf{z}=330$. We trained the model for 200 epochs using ADAM \cite{kingma2014adam} optimizer with a learning rate of $2e^{-4}$ and decay of $0.98$ every epoch. Training took 2.5 h on an A6000 GPU.

\subsection{Atlas Based Brain Segmentation and Age Prediction}
\begin{figure}[t!]
\includegraphics[width=\textwidth]{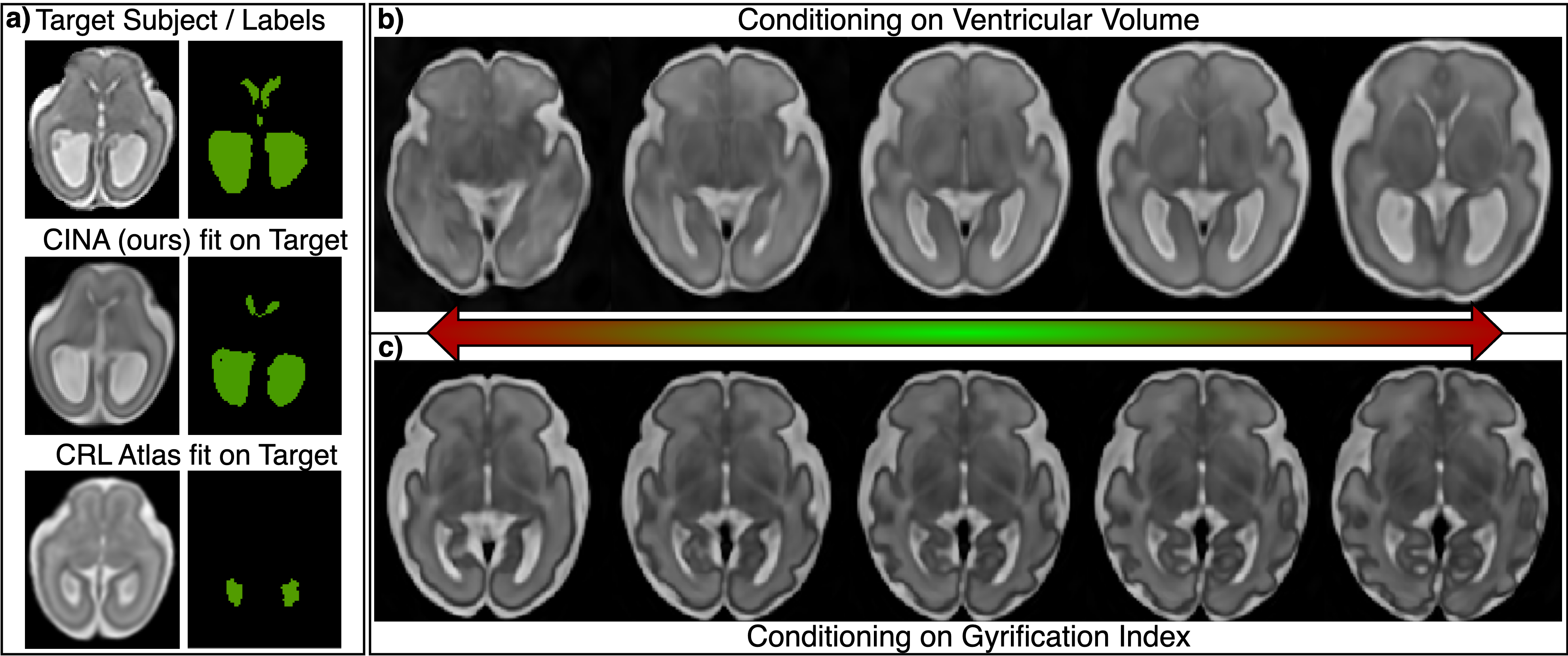}
\centering
\caption{\textbf{a)} CINA and the CRL Atlas are matched with a target subject. The CRL atlas fails to adequately represent the enlarged ventricles. \textbf{b+c)} Interpolation on the conditioned anatomy for a fixed time point. Note, top left and right brains start to show corruptions as these extremes where never encountered during training (extrapolation).} \label{fig:cond_ventricles}
\end{figure}
\begin{figure}[t!]
\includegraphics[width=\textwidth]{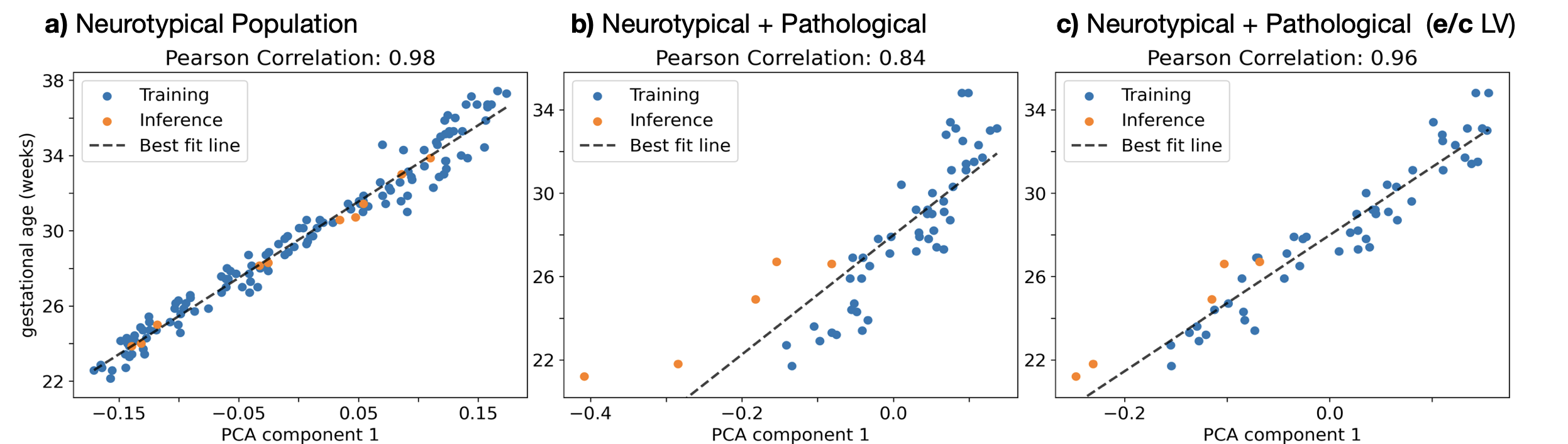}
\caption{Learned representation of gestational age (GA). The first PCA component of a latent code is mapped against the corresponding subject's GA. Blue and orange represent latent codes learned during training and inference. a) A nearly perfect encoding of GA can be observed for the neurotypical population. b) For a mixed population, containing brains with ventriculomegaly, the correlation degrades. c) Explicitly conditioning (\textbf{e/c}) on ventricular volume improves the encoding. Note, in this setup, CINA has never seen a subject's GA.} \label{fig:age_correlation}
\end{figure}

This section evaluates the representation fidelity of CINA and compares it with the fetal brain atlas from the Computational Radiology Laboratory (CRL) at Boston Children’s Hospital~\cite{Gholipour2017}. We test brain age prediction and segmentation. For segmentation, we register the CRL atlas to test subjects using ANTs~\cite{Avants2009}. Next, we project the CRL segmentation on the test brain and compute the dice coefficient between the projected and groundtruth segmentation. CINA does not require a registration. Instead, we get the segmentation by fitting a new latent code $\textbf{z}_t \sim \mathcal{N}(0, 10^{-2})$ to the test subject as described in section \ref{inference} and Fig.~\ref{architecture} c). Note, we only fit the MRI intensities, not the ground truth segmentation. We get the predicted segmentation via Eq.~\ref{eq:forward_pass}. Table \ref{tab:segmentation} presents the results. Both methods show good accuracy across all tissue classes for neurotypical cases cGM represents the most challenging class due to its complex topology. For pathological cases with VM, i.e. brains showing enlarged lateral ventricles (LV), the CRL atlas shows clear deficits for the LV. CINA on the other hand demonstrates better modeling of the LV with a notably higher dice score. Moreover, if we model the LV volume explicitly, denoted as \textbf{e/c}, we observe further improvements. Fig.~\ref{fig:cond_ventricles} a) visualizes a case with VM together with the predictions of CINA and the CRL atlas. Unlike the CRL atlas, CINA adequately captures the LV. Using the CRL atlas we predict a subjects brain age by mapping it to the atlas of similar brain volume and then interpolating the age. For CINA, we regress the age from the learned latent code, utilizing the learned age correlation presented in Fig.~\ref{fig:age_correlation}. 
\begin{table}[htbp]
\centering
\caption{Dice score and mean absolute error of GA (MAE-GA) of CINA and the CRL atlas. Mean values, with standard deviation in parentheses, for four of six brain regions of 10 healthy fetal subjects from dHCP\cite{dhcp_data} and 5 pathological fetal subjects from FeTA with VM\cite{FeTA2021}. \textbf{e/c} denotes explicit conditioning of the LV volume as described in section \ref{anat_conditioning}. $\overline{DSC}$ denotes mean dice of all \textit{six} regions. (All results in Supplementary Table \ref{tab:sup_segmentation}.}
\label{tab:segmentation}

\begin{tabularx}{\textwidth}{@{} l *{6}{X} @{}}
\toprule
Method \quad \quad \quad & CSF & cGM & LV & $\overline{DSC}$ & MAE-GA\\
\midrule
\multicolumn{6}{c}{Neurotypical Brains (dHCP)} \\
\midrule
CRL \cite{Gholipour2017} & 0.83 (0.03) & 0.65 (0.04) & 0.66 (0.06) & 0.76 (0.03) & 1.09 (0.59) \\
CINA & \textbf{0.86 (0.04)} & \textbf{0.69 (0.08)} & \textbf{0.79 (0.06)} & \textbf{0.83 (0.04)} & \textbf{0.23 (0.21)} \\
\midrule
\multicolumn{6}{c}{Pathological Brains with Ventriculomegaly (FeTA)} \\
\midrule
CRL & 0.74 (0.05) & 0.43 (0.10) & 0.41 (0.16) & 0.60 (0.12) & \textbf{1.14 (0.82)}\\
CINA & 0.77 (0.08) & 0.50 (0.14) & 0.73 (0.20) & 0.67 (0.18) & 2.61 (1.34)\\
CINA (\textbf{e/c} LV) & \textbf{0.79 (0.08)} & \textbf{0.52 (0.21)} & \textbf{0.81 (0.07)} & \textbf{0.70 (0.18)} & 1.24, (0.44)\\
\bottomrule
\end{tabularx}
\end{table}

\subsection{Atlas Construction}
\begin{figure}[t!]
\includegraphics[width=\textwidth]{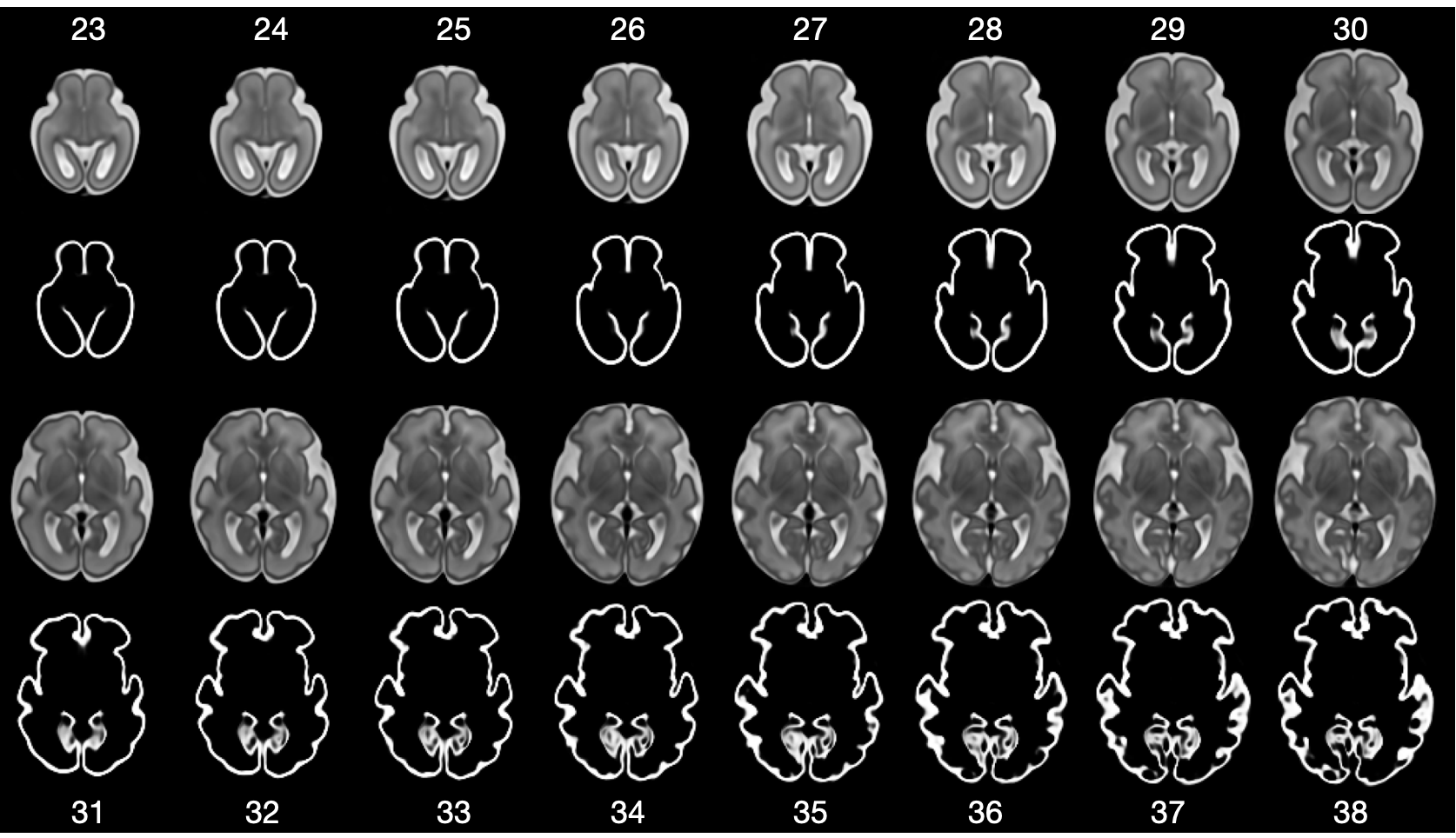}
\caption{Temporal fetal brain atlas from 23 to 38 GA with corresponding probability maps of the cortical grey matter.} \label{fig:temporal_atlas}
\end{figure}
CINA implicitly represents a spatial- and time-continuous atlas. Contrary to 3D image based atlases that are only available for fixed, discrete time points and resolution, CINA can generate atlases for any time point and resolution. Fig.~\ref{fig:temporal_atlas} presents a temporal atlas generated with CINA for GA 23 to 38. Moreover, CINA can generate the temporal atlases while simultaneously conditioning on learned anatomical characteristics, visualized by Fig.~\ref{fig:cond_ventricles}~b+c) for LV volume and Giryfication Index of the cortex for a specific time point.    

\section{Conclusion}
We have introduced CINA, a novel conditional implicit neural atlas designed for the continuous spatial and temporal generation of 3D fetal brain atlases. CINA can model key anatomical features, including ventricular volume or degree of cortical folding. This facilitates accurate representation of neurotypical brains, but also of pathological cases where classical atlases often fall short. Moreover, CINA achieves a high compression rate, encapsulating the representation of hundreds of brain states across varying resolutions, gestational age, and anatomical expressions, within a mere 30 MB footprint. This compact yet versatile format enables end-users to generate detailed brain atlases in real-time, tailored to their specific needs.

\begin{credits}
\subsubsection{\ackname} This study was supported by the ERC (Deep4MI - 884622), and by the ERA-NET NEURON Cofund (MULTI-FACT - 8810003808).
\end{credits}

%
%
%
\bibliographystyle{splncs04}
\bibliography{Reference}
\clearpage
\section{Supplementary Material}
\begin{figure}
\includegraphics[width=\textwidth]{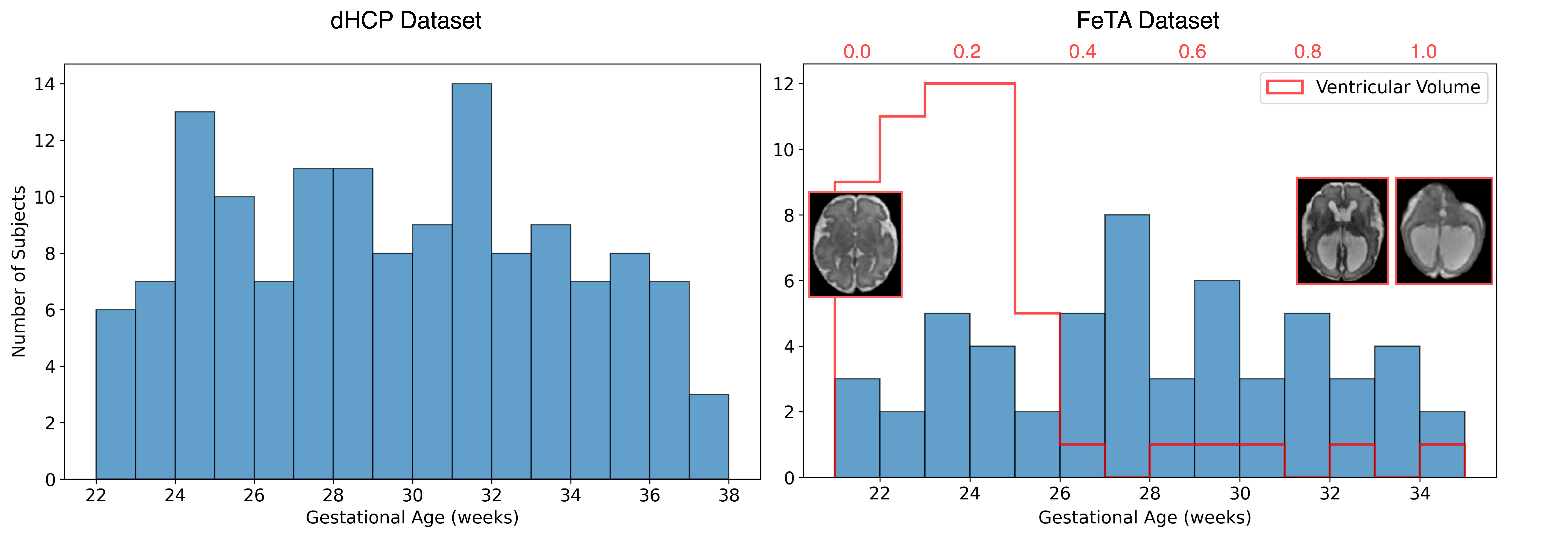}
\caption{Distribution of gestational age of the selected fetal subjects for the dHCP (left) and FeTA (right) dataset. The right histogram additionally shows the distribution (red) of the subjects by ventricular volume, normalized between 0 and 1.} 
\label{fig:histogram}
\end{figure}

\begin{figure}
\includegraphics[width=\textwidth]{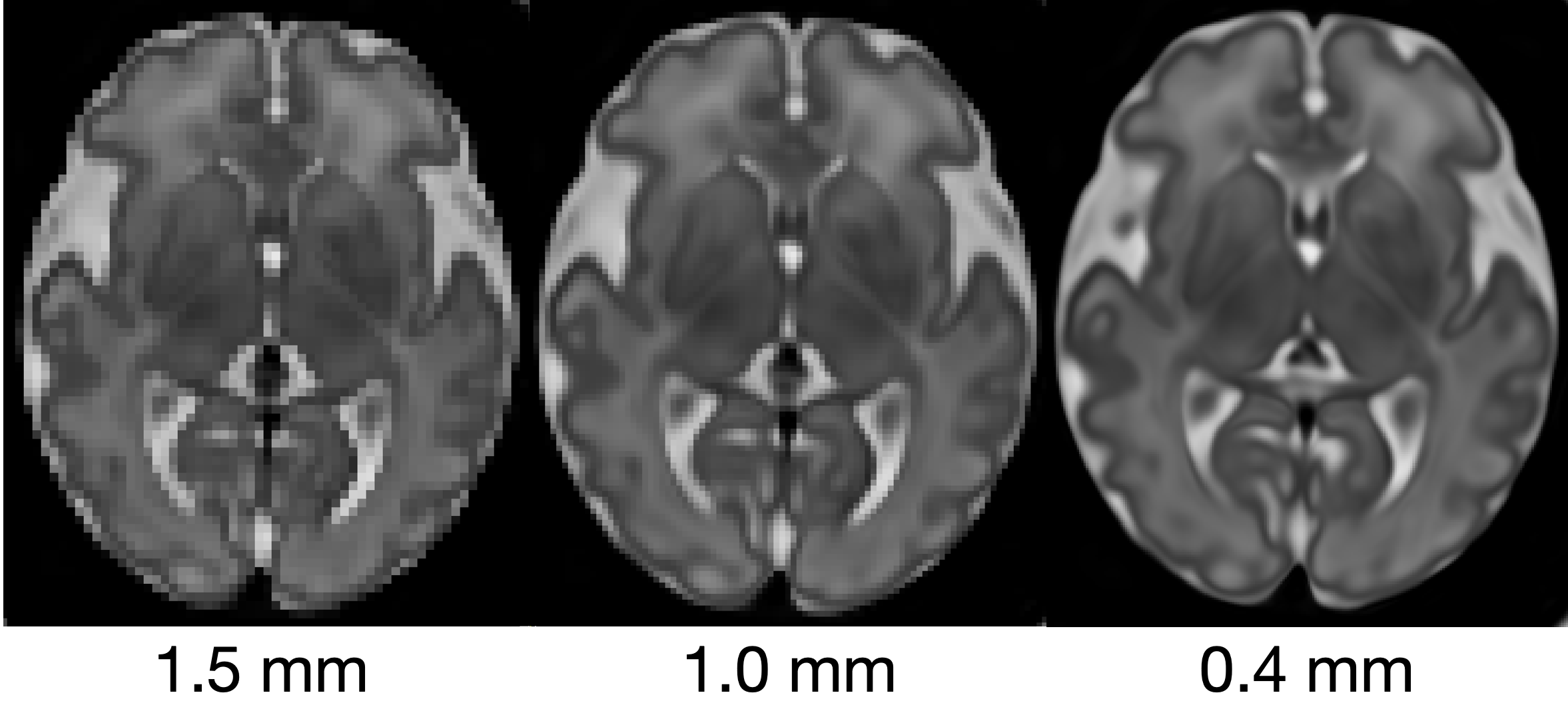}
\caption{CINA is resolution agnostics allowing us to generate an atlas of arbitrary spatial resolution. Left to right shows an atlas for a specific time point of increasing spatial resolutions of 1.5mm, 1.0mm, and 0.4mm isotropic voxel spacing.  } 
\label{fig:resolution}
\end{figure}

\begin{table}[htbp]
\centering
\caption{Dice score and mean absolute error of GA (MAE-GA) of CINA and the CRL atlas. Mean values, with standard deviation in parentheses, for four of six brain regions of 10 neurotypical fetal subjects from dHCP\cite{dhcp_data}, 5 pathological fetal subjects with ventriculomegaly from FeTA and 5 neurotypical and pathological brains with no or only mild VM \cite{FeTA2021}. \textbf{e/c} denotes the setup with explicit conditioning of the lateral ventricular volume as described in section \ref{anat_conditioning}. $\overline{DSC}$ denotes mean dice over all six regions.}
\label{tab:sup_segmentation}

\begin{tabularx}{\textwidth}{@{} l *{6}{X} @{}}
\toprule
Method \quad \quad \quad & CSF & cGM & LV & $\overline{DSC}$ & MAE-GA\\
\midrule
\multicolumn{6}{c}{Neurotypical Brains (dHCP)} \\
\midrule
CRL \cite{Gholipour2017} & 0.83 (0.03) & 0.65 (0.04) & 0.66 (0.06) & 0.76 (0.03) & 1.09 (0.59) \\
CINA & \textbf{0.86 (0.04)} & \textbf{0.69 (0.08)} & \textbf{0.79 (0.06)} & \textbf{0.83 (0.04)} & \textbf{0.23 (0.21)} \\
\midrule
\multicolumn{6}{c}{Pathological Brains with Ventriculomegaly (FeTA)} \\
\midrule
CRL & 0.74 (0.05) & 0.43 (0.10) & 0.41 (0.16) & 0.60 (0.12) & \textbf{1.14 (0.82)}\\
CINA & 0.77 (0.08) & 0.50 (0.14) & 0.73 (0.20) & 0.67 (0.18) & 2.61 (1.34)\\
CINA (\textbf{e/c} LV) & \textbf{0.79 (0.08)} & \textbf{0.52 (0.21)} & \textbf{0.81 (0.07)} & \textbf{0.70 (0.18)} & 1.24, (0.44)\\
\midrule
\multicolumn{6}{c}{Neurotypical Brains and Pathological Brains without VM (FeTA)} \\
\midrule
CRL & 0.80 (0.03) & \textbf{0.56 (0.08)} & 0.64 (0.04) & 0.72 (0.09) & 0.81 (0.41)\\
CINA & \textbf{0.83 (0.05)} & 0.50 (0.10) & 0.75 (0.04) & \textbf{0.74 (0.11)} & \textbf{0.50 (0.44)}\\
CINA (\textbf{e/c} LV) & \textbf{0.83 (0.05)} & 0.49 (0.10) & \textbf{0.77 (0.04)} & 0.73 (0.12) & 0.85 (0.72)\\
\bottomrule
\end{tabularx}
\end{table}

\end{document}